\documentclass{article}

\usepackage[preprint, nonatbib]{neurips_2026}

\usepackage[utf8]{inputenc} \usepackage[T1]{fontenc}    \usepackage{hyperref}       \usepackage{url}            \usepackage{booktabs}       \usepackage{amsfonts}       \usepackage{nicefrac}       \usepackage{microtype}      \usepackage{xcolor}         

\usepackage{graphicx}
\usepackage{subcaption}

\usepackage{amsmath}
\usepackage{amssymb}
\usepackage{mathtools}
\usepackage{amsthm}
\usepackage{algorithm}
\usepackage{algpseudocode}

\usepackage[numbers]{natbib}

\title{Phasor Memory Networks: Stable Backpropagation Through Time for Scalable Explicit Memory}

\author{Sungwoo~Goo \\
  College of Pharmacy\\
  Chungnam National University\\
  Daejeon, South Korea \\
  \texttt{swgoo91@gmail.com} \\
\And
  Hwi-yeol Yun \\
  College of Pharmacy\\
  Chungnam National University\\
  Daejeon, South Korea \\
  \texttt{hyyun@cnu.ac.kr} \\
  \AND
  Sangkeun Jung \\
  Department of Computer Science \& Engineering\\
  Chungnam National University\\
  Daejeon, South Korea \\
  \texttt{hugmanskj@gmail.com} \\
}

\begin{document}

\graphicspath{{./fig_img/}}

\maketitle

\begin{abstract}
  For over a decade, explicit memory architectures like the Neural Turing Machine have remained theoretically appealing yet practically intractable for language modeling due to catastrophic gradient instability during Backpropagation Through Time.
In this work, we break this stalemate with \textit{Phasor Memory Network} (PMNet), a novel architecture that structurally resolves memory volatility through \textit{Unitary Phasor Dynamics} and \textit{Hierarchical Learnable Anchors}.
Rather than relying on brute-force scaling, we present a mechanistic proof-of-concept in a controlled byte-level setting.
By constraining recurrent state updates to phase rotations on a complex unit circle, PMNet preserves gradient norms and inherently prevents divergence without the need for specialized initialization.
We empirically demonstrate the active actuation of the memory module through a synthetic Copy-Paste task,
where PMNet utilizes an expansive \textit{85-slot hierarchical memory tree} ($=\sum^{4}_{h=1}4^{h-1}$) to achieve near 100\% exact retrieval across temporal distances that completely exceed the local sliding window attention's receptive field.
Furthermore, despite being a compact 119M parameter model trained on 18.8B tokens, PMNet matches the zero-shot long-context robustness of a Mamba model that is three times larger.
Our ablation studies and gradient analyses confirm that the historical failure of explicit memory was a structural alignment problem, which PMNet effectively overcomes, providing a theoretically grounded foundation for scalable sequence modeling. \end{abstract}

\begin{figure*}[tbhp]
    \centering
    \includegraphics[width=.85\textwidth]{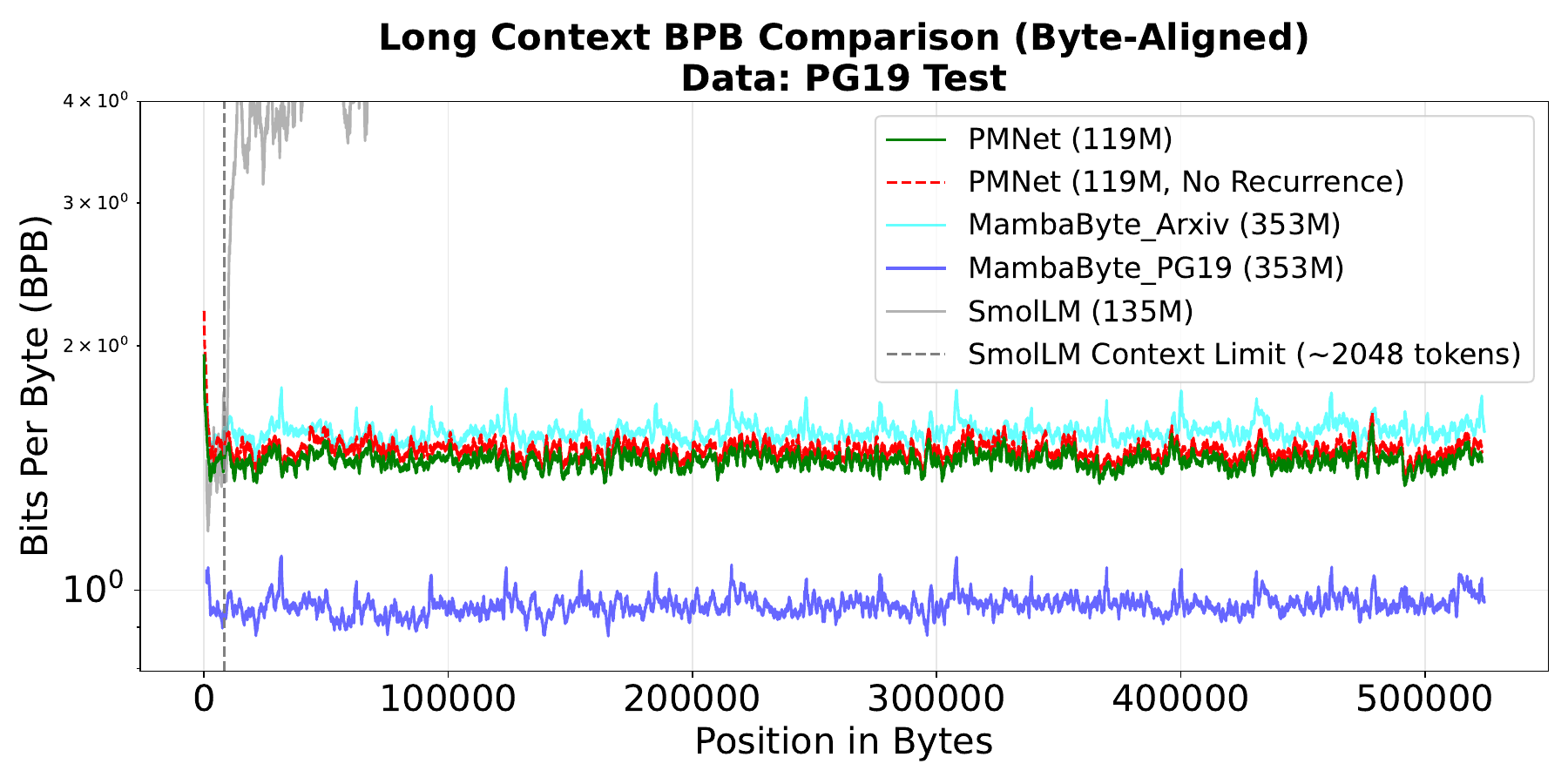} 
    \caption{\textbf{Evaluation of Zero-shot Generalization on PG-19.} 
    Comparison of Byte-level Perplexity (BPB, lower is better) on the PG-19 test set. 
    \textbf{(Green) PMNet (119M) (Ours):} Trained on FineWeb-Edu. Despite the \textit{dual disadvantage} of being $3\times$ smaller than the baseline and evaluated in a zero-shot setting, PMNet maintains stable extrapolation up to 512k bytes. This confirms that our Phasor Dynamics capture universal linguistic structures irrespective of domain training. 
    \textbf{(Cyan) MambaByte\_Arxiv (353M):} An OOD baseline. PMNet achieves parity with this significantly larger model, demonstrating superior parameter efficiency. 
    \textbf{(Blue) MambaByte\_PG19 (353M):} Shown as an in-domain oracle reference.
    \textbf{(Red Dashed) Ablation:} Performance degrades without memory updates, proving that the stability stems from the memory mechanism, not the SWA. 
    \textbf{(Gray) SmolLM (135M):} Demonstrates the catastrophic failure of standard Transformers beyond their context window ($\approx 2k$), contrasting sharply with PMNet's linear stability.}
    \label{fig:long_context}
\end{figure*} 
\section{Introduction}
The recent evolution of Large Language Models (LLMs) has fundamentally transformed the landscape of artificial intelligence.
However, as the demand for processing massive documents and long-form reasoning grows,
the standard Transformer architecture faces a critical bottleneck: the quadratic computational complexity of its self-attention mechanism.
This $O(L^2)$ scaling makes it prohibitively expensive to maintain ultra-long context windows.
To address this, alternative architectures such as Recurrent Neural Networks (RNNs) and State Space Models (SSMs) like Mamba have been proposed to achieve linear-time complexity.
Despite their inference efficiency, training these models on extremely long sequences remains a formidable challenge.
Rather than inherent flaws in gating mechanisms,
the primary difficulties arise from managing non-unitary state dynamics and the accumulation of numerical errors during Backpropagation Through Time (BPTT).
Without strict structural constraints, the recurrent states in these linear models are prone to unbounded growth or catastrophic decay.
Consequently, modern linear architectures often rely on meticulous initialization or forced exponential decay to prevent gradients from exploding,
which inherently limits their ability to flawlessly retrieve exact information across extended temporal horizons.

In this paper, we argue that stabilizing BPTT requires a fundamental shift in learning mechanics rather than heuristic optimization patches.
We introduce \textbf{Phasor Memory Network~(PMNet)}, a novel architecture that provides a mechanistic proof-of-concept for stable, linear-time explicit memory.
PMNet goes beyond merely storing hidden states; it explicitly separates static semantic anchors (memory embeddings) from dynamic context recurrence, allowing the model to directly access memory slots without catastrophic addressing drift.
PMNet is built upon three core technical innovations:

\textbf{Phasor Dynamics for Unitary Recurrence} \\
We address the spectral instability of recurrent states by encoding information as phase angles on a complex unit circle.
This enables purely additive updates that are structurally bounded and unitary,
ensuring numerical stability and strictly preserving gradient norms without vanishing or exploding gradients,
even over ultra-long sequences.

\textbf{Hierarchical Memory with Learnable Embeddings} \\
Because Phasor Dynamics guarantee stable BPTT,
we can afford to implement a massive, tree-structured explicit memory hierarchy without the risk of gradient collapse.
Each node is associated with a learnable embedding that acts as a semantic anchor,
enabling efficient $O(1)$ routing.
Unlike traditional Neural Turing Machines (NTMs) \cite{graves2014neural} that suffer from volatile content-based addressing, these learnable anchors provide structurally permanent reference points.
This guarantees that historical knowledge remains reliably retrievable,
while the phasor states handle the continuous injection of new context.

\textbf{Parallel Segmented Scan \& Adaptive Scaling} \\
To prevent gradient explosions caused by imbalanced routing collisions within the tree structure,
we implement a parallelized segmented scan mechanism.
By introducing segment-aware gradient normalization based on update frequency,
we ensure robust optimization and strictly bounded energy flow across the memory hierarchy.

Crucially, we validate the necessity and efficacy of Phasor Dynamics through rigorous empirical analysis of learning dynamics.
Rather than solely relying on final perplexity scores, we analyze the training gradient norms to reveal a clear phase transition in the model's behavior.
We observed that local Sliding Window Attention (SWA) dominates early optimization, but as the required lookback distance exceeds the SWA's receptive field, the global Phasor Memory actively awakens to route and retrieve exact information.
Leveraging this structurally guaranteed stability,
we adopt a \textbf{byte-level} modeling approach to eliminate tokenizer bias.
Although raw byte sequences result in significantly longer contexts that typically cripple standard models,
PMNet flawlessly executes a synthetic associative recall (Copy-Paste) task over thousands of steps.
Furthermore, when scaled to 18.8B training tokens,
a compact 119M parameter PMNet matches the zero-shot long-context robustness of an in-domain MambaByte model three times its size.
We provide the source code at our repository\footnote{\url{https://github.com/swgoo/pmnet}}. 
\section{Related Works}
The pursuit of efficient long-context modeling has led to several distinct research trajectories, ranging from linearizing attention to reviving recurrent dynamics and external memory systems.
PMNet stands at the intersection of these fields, addressing the fundamental trade-offs between computational efficiency and stable information retention.

\subsection{Linear Transformers and Recurrent Models}
To overcome the $O(L^2)$ complexity of standard Transformers, various linear-time architectures have emerged. 
Linear Transformers~\cite{katharopoulos2020transformers} and models like RWKV~\cite{peng2023rwkv} and RetNet~\cite{sun2023retentive} reformulate attention into a dual form that allows for recurrent inference. 
While these models achieve $O(L)$ efficiency, they frequently rely on sophisticated, magnitude-based gating mechanisms to prevent state collapse. 
PMNet diverges from this trend by eliminating traditional gates entirely. 
Instead, it utilizes Phasor Dynamics to ensure state stability through rotation on a complex unit circle, providing a structural guarantee against the vanishing gradient issues often found in purely additive recurrent structures.

\subsection{Long-context Transformers.} 
Several approaches have been proposed to extend the context window of Transformers.
\citet{dai2019transformer} introduced segment-level recurrence (Transformer-XL) to capture longer dependencies, while \citet{press2021train} proposed ALiBi to enable length extrapolation beyond the training horizon.
However, these methods still rely on attention mechanisms with quadratic or linear-growing complexity.
In the realm of explicit memory, \citet{wu2022memorizing} utilized k-NN based lookup tables (Memorizing Transformers). 
Unlike these approaches, PMNet employs a fully differentiable, hierarchical phasor memory that achieves $O(1)$ inference complexity with structurally bounded stability.

\subsection{State Space Models (SSMs)}
Recently, SSMs such as S4~\cite{gu2021efficiently} and Mamba~\cite{gu2024mamba} have demonstrated state-of-the-art performance in long-sequence tasks by leveraging discretized linear systems.
However, the performance of SSMs is often highly sensitive to initialization and the scaling of the time-step parameter ($\Delta$), which can lead to numerical instability during large-scale training~\cite{smith2022simplified}.
PMNet offers a more robust alternative; by constraining state updates to a unitary manifold, it achieves superior stability using standard initialization techniques. 
This inherently bounded dynamics significantly lowers the engineering overhead compared to the meticulously tuned spectral constraints required by modern SSMs.

\subsection{Conditional Memory and Knowledge Lookup}
The paradigm of augmenting neural networks with external memory, originally pioneered by NTMs~\cite{graves2014neural, graves2016hybrid}, has recently witnessed a resurgence under the framework of Conditional Memory~\cite{cheng2026conditional}.
Current research posits that standard Transformers lack a native primitive for efficient knowledge lookup, thereby compelling the model to simulate retrieval through computationally expensive neural operations.
To bridge this gap, architectures such as Engram have introduced static embedding lookups to offload routine reconstruction tasks.
PMNet translates this philosophy into a linear recurrent framework by implementing a Hierarchical Memory with Learnable Embeddings.
Unlike traditional NTMs, which are prone to addressing drift driven by volatile content-based mechanisms, PMNet employs a fixed, tree-structured hierarchy anchored by learnable embeddings.

\subsection{Unitary Dynamics and Rotational Embeddings}
Our work also draws inspiration from Unitary RNNs and Rotary Positional Embeddings (RoPE), which utilize complex-valued representations to preserve norms~\cite{trabelsi2017deep, arjovsky2016unitary,su2021roformer}.
While RoPE applies rotation to individual token embeddings to encode relative positions, PMNet applies rotational dynamics to the state update rule itself. 
This ensures that the cumulative memory of the model remains strictly bounded and unitary over time. 
By integrating this unitary constraint with a hierarchical lookup mechanism, PMNet provides a unique structural guarantee against numerical divergence, enabling stable processing of contexts up to 512k tokens. 
\section{Methods}
\begin{figure}[tbhp]
    \centering
    \includegraphics[width=.78\textwidth]{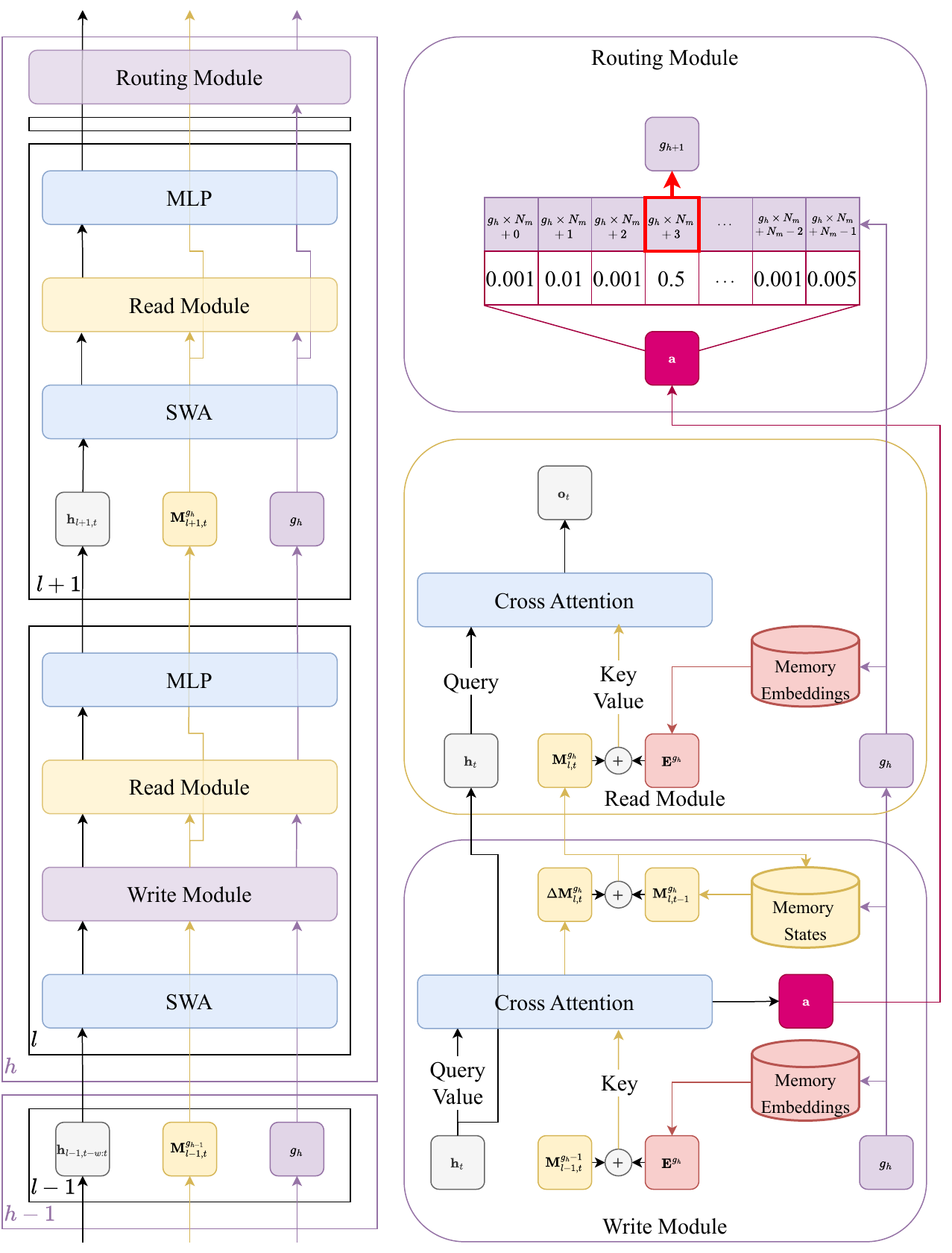} 
    \caption{
    \textbf{Overview of the PMNet Architecture.} \\
    The model integrates local processing with a global hierarchical memory system.
    The global context is organized as a sparse tree structure. At each hierarchy level $h$, a routing mechanism selects a specific active memory group index $g_h$, allowing the total memory capacity to scale exponentially with depth while maintaining $\mathcal{O}(1)$ access complexity per layer.}
\label{fig:arch}
\end{figure} 
The PMNet architecture addresses the scalability limitations of standard Transformers by decoupling local processing from global context management. 
Based on the observation that attention density is heavily concentrated locally, we employ Sliding SWA \cite{child2019generating} for short-range dependencies and a novel Hierarchical Phasor Memory for global context retention.
PMNet organizes memory as a deep, sparse tree. 
Furthermore, we mitigate the gradient instability inherent in RNNs by utilizing unitary phasor dynamics.
This ensures stable BPTT over sequences exceeding 8k tokens by maintaining the gradient norm via rotation in the complex plane, rather than scaling magnitude. 
For brevity, we omit the layer index $l$, time step $t$, Memory group index $g$ , memory hierarchy level $h$, and batch dimension in the following equations unless necessary for clarity. All operations are performed element-wise or broadcasted across the batch dimension appropriately.

Table \ref{tab:notations} summarizes the key notations and tensor shapes used in this section.

\subsection{Hierarchical Phasor Memory}
We consider a transformer backbone consisting of $L$ layers. 
To efficiently capture long-range dependencies, PMNet augments this backbone with a hierarchical external memory that is updated periodically every $P$ layers.
This design induces a logical hierarchy of depth $H = L/P$. 
Let $\mathbf{h}_{l,t} \in \mathbb{R}^d$ denote the hidden state at layer $l$ and time step $t$.
The memory is organized as a tree-structured addressing space where each node represents a discrete memory group. 
At each hierarchy level $h \in \{1, \dots, H\}$, a group is defined by a set of $N_m$ phasor vectors, denoted as $\mathbf{M}_t^{g_h} \in [-\pi, \pi]^{N_m \times d_m}$, where $d_m$ is the memory dimension and $g_h$ is the active memory group index. 
Crucially, the tree structure allows the total memory capacity $C_m$ to scale exponentially (${N_m}^H$) while ensuring that each token only traverses a sparse, $\mathcal{O}(1)$ routing path. 
By constraining all updates to the phasor domain $[-\pi, \pi]$, we structurally guarantee that the total energy of the memory remains bounded, preventing numerical divergence. 

\subsubsection{Sparse Tree Routing}
To maintain logarithmic access complexity, PMNet does not access the global memory bank simultaneously.
Instead, the model traverses a single path through the $N_m$-ary tree structure.
The routing decision for the subsequent level $h+1$ is determined by a content-aware alignment between the current hidden state and the phase-domain memory state.
First, we project the hidden state into a query phase vector $\phi_q$ using a hyperbolic tangent scaled by $\pi$:
\begin{equation}
    \phi_q = \pi \tanh(\mathbf{W}_q \cdot \text{RMSNorm}(\mathbf{h}))
\end{equation}

To interact with the angular memory state $\mathbf{M}$, we employ a geometric projection $\Phi(\cdot)$ that preserves cyclic continuity. 
We also introduce Learnable Memory Embeddings $\mathbf{E} \in \mathbb{R}^{N_m \times d_m}$, which serve as semantic anchors to stabilize addressing:
\begin{equation}
    \Phi(\theta) = [\sin(\theta), \cos(\theta)] \qquad
    \phi_k = \pi \tanh\left(\mathbf{W}_k \cdot \Phi(\mathbf{M}_{t}^{g_h} + \mathbf{E}^{g_h})\right)
\end{equation}

The routing distribution $\mathbf{a} \in \mathbb{R}^{N_m}$ is computed via cosine similarity in the phase domain. Specifically, we calculate the cosine of the phase difference element-wise and sum these values across the feature dimension $d_{mem}$:
\begin{equation}
    \mathbf{a} = \text{softmax}\left( \frac{\sum_{j=1}^{d_{mem}}\cos(\phi_{q,j} - \phi_{k,j})}{\sqrt{d_{mem}}} \right)
\end{equation}
Simultaneously, the specific memory group index $g$ for the next hierarchy level is selected via top-1 selection (argmax). This creates a dynamic, token-specific path through the memory tree:
\begin{equation}
    g_{h+1} = g_h \cdot N_m + \operatorname*{argmax}(\mathbf{a}), \quad \text{with } g_0 = 0
\end{equation}
where $g_h$ denotes the memory group index at hierarchy level $h$. The root index $g_0$ is initialized to $0$ for all inputs, ensuring a unified starting point for the hierarchical addressing.

\subsubsection{Unitary Phasor Update}
Standard RNNs struggle with long-term dependencies due to the spectral radius of the recurrent matrix deviating from $1$. 
PMNet enforces unitarity by defining the memory update strictly as a rotation in the complex plane, implemented via modulo arithmetic. 

The update content $\Delta \mathbf{M}$ is derived from the value projection of the hidden state:
\begin{equation}
    \mathbf{v} = \mathbf{W}_v {\mathbf{h}}, \quad \mathbf{m} = \pi \tanh(\mathbf{W}_{out} \mathbf{v}) \qquad
    \Delta \mathbf{M} = \mathbf{a} \cdot \mathbf{m}^\top
\end{equation}

The new memory state is updated additively modulo $2\pi$:
\begin{equation}
    \mathbf{M}_{t} = (\mathbf{M}_{t-1} + \Delta \mathbf{M}) \pmod{2\pi}
\end{equation}
This formulation is theoretically equivalent to the unitary update $z_t = z_{t-1} \cdot e^{i\Delta \mathbf{M}}$, ensuring the gradient norm remains invariant ($|e^{i\theta}| \equiv 1$) regardless of sequence length.

\subsubsection{Memory Read via Phase Cross-Attention}
To retrieve global context from the updated memory, PMNet employs a specialized cross-attention mechanism where the current hidden state acts as the query and the active memory group serves as the key-value store.
Consistent with the routing and update dynamics, similarity is computed in the phase domain to preserve the geometric properties of the memory.
Let $\tilde{\mathbf{M}} = \mathbf{M}_t^{g_h} + \mathbf{E}^{g_h}$ denote the embedding-augmented memory state.
We derive the query phase $\phi_q$ from the hidden state, and the key phase $\phi_k$ and value vectors $\mathbf{V}$ from the geometric projection of the memory:
\begin{equation}
[\mathbf{K} ; \mathbf{V}] = \mathbf{W}_{kv} \Phi(\tilde{\mathbf{M}}) \qquad
\phi_q = \pi \tanh(\mathbf{W}_q \mathbf{h}), \quad \phi_k = \pi \tanh(\mathbf{K})
\end{equation}

The retrieved context $\mathbf{o}_t$ is the result of aggregating the memory values weighted by their phase alignment with the query, allowing the model to selectively access relevant historical information:
\begin{equation}
\mathbf{o}_t = \mathbf{W}_o \sum_{i=1}^{N_m} \text{softmax}\left( \frac{\sum_{j} \cos(\phi_{q,j} - \phi_{k,i,j})}{\sqrt{d_{mem}}} \right)_i \mathbf{V}_i
\end{equation}
\subsection{Gradient Stabilization via Segment-Aware Normalization}
While the periodic activation function $\pmod{2\pi}$ prevents value explosion in the forward pass, the additive nature of memory updates can lead to gradient explosion during backpropagation, especially when multiple tokens update the same memory slot simultaneously. 
To address this, we introduce Segment-Aware Gradient Normalization. 
Let $s$ be the collision count, defined as the number of tokens in the current sequence that write to the specific memory group $g_h$. 
This effectively represents the length of the continuous update segment for that memory slot.
To stabilize the learning dynamics without constraining the representational power of the forward memory state, we employ a pseudo-normalization technique using the stop-gradient operator ($\text{sg}[\cdot]$):\begin{equation}\Delta \tilde{\mathbf{M}} = \frac{\Delta \mathbf{M}}{\sqrt{s}} + \text{sg}\left[\Delta \mathbf{M} - \frac{\Delta \mathbf{M}}{\sqrt{s}}\right]\end{equation}In the forward pass, this simplifies to $\Delta \tilde{\mathbf{M}} = \Delta \mathbf{M}$, allowing the memory to accumulate full phase information. 
However, during backpropagation, the gradient is scaled by $1/\sqrt{s}$:
$\frac{\partial \mathcal{L}}{\partial \Delta \mathbf{M}} = \frac{1}{\sqrt{s}} \cdot \frac{\partial \mathcal{L}}{\partial \Delta \tilde{\mathbf{M}}}$
This ensures that the variance of the accumulated gradients over a segment of length $s$ approximates $\text{Var}(\sum_{i=1}^s \nabla \mathbf{M}) \approx s \cdot (1/\sqrt{s})^2 \times \text{Var}(\nabla \mathbf{M}) = \text{Var}(\nabla \mathbf{M})$, effectively preventing gradient explosion for frequently accessed memory slots while maintaining unit variance for the learning signal.
The detailed hardware-aware implementation of the segmented scan is provided in Algorithm \ref{alg:segmented_scan} of the Appendix. 
\section{Experiments}
In this section, we empirically validate the learning mechanics and structural advantages of PMNet.
We design our evaluation to answer three fundamental questions:
(1) Does explicit memory provide an architectural advantage over linear SSMs under strictly controlled conditions?
(2) \textit{Why} are unitary phasor dynamics necessary for BPTT, and how does the memory module awake during training?
(3) Does this structurally guaranteed stability scale to ultra-long contexts and match larger foundational models?

\begin{figure*}[tb]
    \centering
    \includegraphics[width=.95\textwidth]{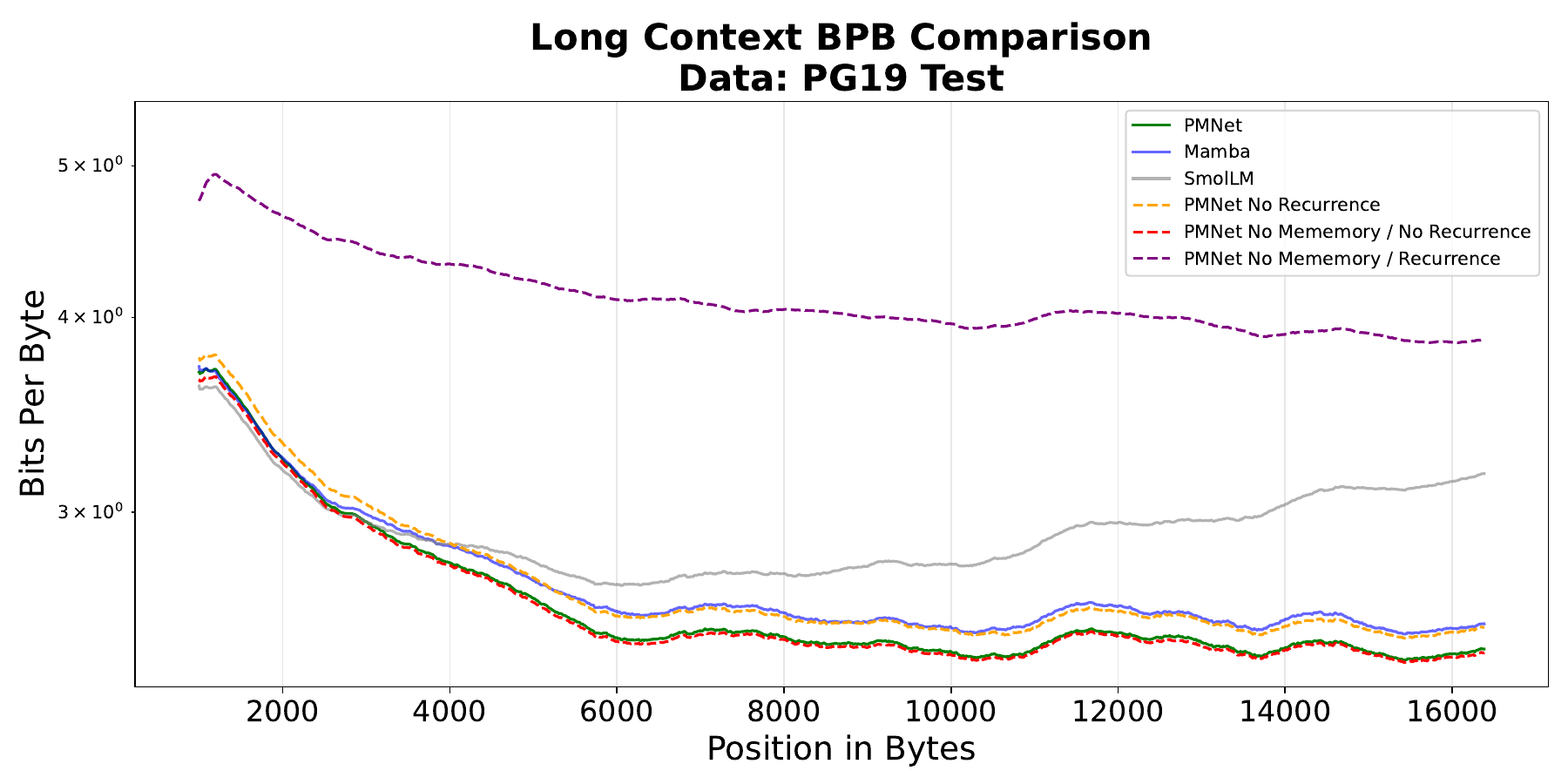} 
    \caption{\textbf{Controlled Architectural Comparison.} Evaluation of $\approx$30M parameter models trained from scratch on 0.5B FineWeb-Edu tokens.
    PMNet consistently outperforms parameter-matched SmolLM and Mamba baselines.
    We ablate memory and recurrence: \textit{No Recurrence} disables recurrence only during evaluation; \textit{No Memory / No Recurrence} serves as a pure SWA baseline (disabled in both phases); and \textit{No Memory / Recurrence} disables memory during training but retains the recurrent evaluation pathway. Results demonstrate PMNet's superior long-range compression and the data-hungry nature of global explicit memory at this scale.}
    \label{fig:controlled_comparison}
\end{figure*} \subsection{Controlled Architectural Comparison}

\textbf{Experimental Setup:}
To rigorously isolate the architectural efficiency of PMNet from confounding factors such as pre-training data mixtures and scale,
we established a strictly controlled experimental setup.
We trained models from scratch with approximately \textit{30M parameters} on an identical subset of exactly \textit{0.5B tokens} sampled from the FineWeb-Edu dataset \cite{penedo2024fineweb}.
All models operate directly at the byte level to test universal sequence modeling capabilities without subword tokenizer biases.
We enforced a strict context length of 2k bytes per sample and utilized identical training recipes across all runs,
including the AdamW optimizer \cite{loshchilov2017decoupled}, cosine learning rate decay, and matched batch sizes.

\textbf{Results \& Analysis:}
As illustrated in Figure \ref{fig:controlled_comparison}, PMNet consistently outperforms both the SmolLM (Transformer) and Mamba baselines under these strictly matched conditions.
Specifically, PMNet achieves a lower Bits per Byte (BPB) in long-context scenarios, exhibiting superior long-range dependency modeling capabilities.
The architectural implications are twofold.
First, compared to SmolLM, PMNet demonstrates that augmenting a local SWA with a global, explicit Phasor Memory yields superior contextual compression without the quadratic scaling bottleneck.
Second, compared to Mamba---which relies on a single continuous hidden state forced into exponential decay (lossy compression)---PMNet's multi-slot explicit memory and unitary phasor updates successfully preserve high-resolution information over thousands of steps without memory degradation.

\textbf{Ablation Study \& Simplicity Bias:}
Our ablation analysis reveals critical insights into the learning dynamics of explicit memory. 
The \textit{PMNet (No Memory / Recurrence)} ablation---which disables memory updates during training but retains the recurrent evaluation pathway---performs strictly worse than the full PMNet, empirically confirming that explicit memory routing is actively learned. 
Interestingly, the pure SWA baseline (\textit{No Memory / No Recurrence}) achieves performance comparable to the full model under this severely constrained 0.5B data regime.
We attribute this parity to the \textbf{simplicity bias} of gradient descent: at a limited data scale, local SWA optimization dominates the early learning phase because short-range patterns are easier to capture \cite{simon2026there}.
The complex global routing schema of explicit memory, conversely, is inherently ``data-hungry'' and requires a substantially larger corpus of long-range dependencies to fully saturate its capacity (as demonstrated later in Section \ref{sec:zero_shot_extrapolation}).

\subsection{Mechanistic Probe: Copy-Paste Task}
\begin{figure}[tbhp]
    \centering
    \includegraphics[width=.75\columnwidth]{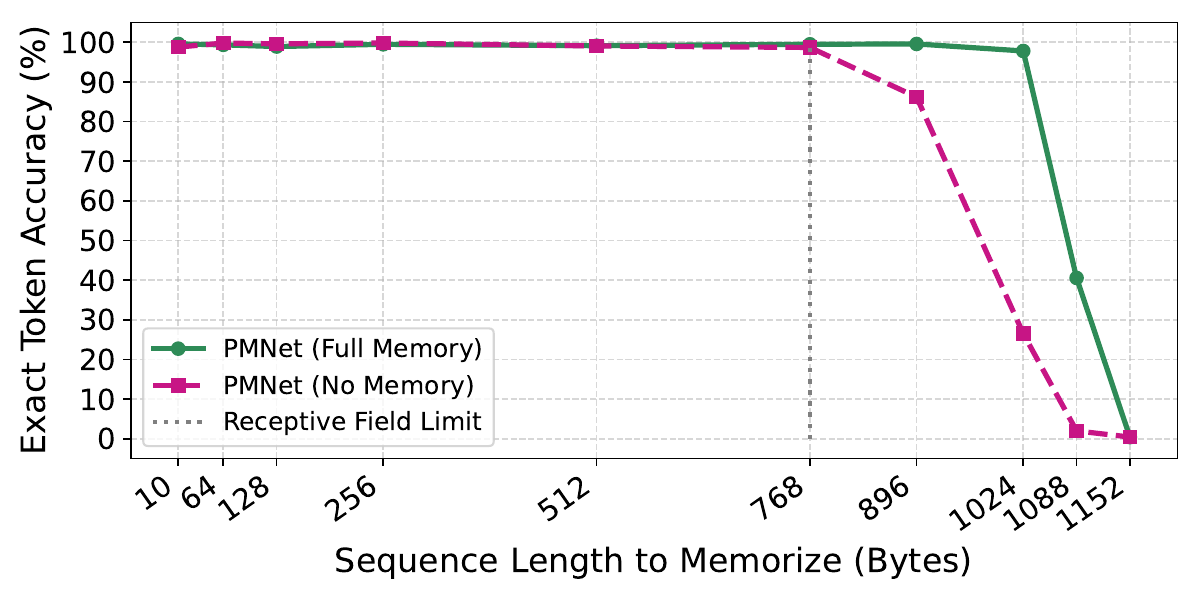}
    \caption{\textbf{Mechanistic Validation via Copy-Paste Accuracy.}
    The SWA-only baseline (\textit{No Memory}) exhibits a catastrophic collapse in accuracy at sequence lengths $N > 768$ as the required lookback distance ($2N$) exceeds its receptive field limit of 1,536 bytes ($12 \text{ layers} \times 128 \text{ window size}$).
    In contrast, despite being trained on a variable distribution of lengths $N \in [10, 1024]$—corresponding to a maximum lookback distance of 2,048 bytes—PMNet maintains near-perfect accuracy across the entire range.
    These results provide definitive proof that the explicit memory module effectively routes and retrieves exact information from temporal horizons that the local attention backbone cannot bridge.}
    \label{fig:copy_paste_accuracy}
\end{figure} \begin{figure}[tbhp]
    \centering
    \begin{subfigure}[t]{0.45\textwidth}
        \centering
        \includegraphics[width=\textwidth]{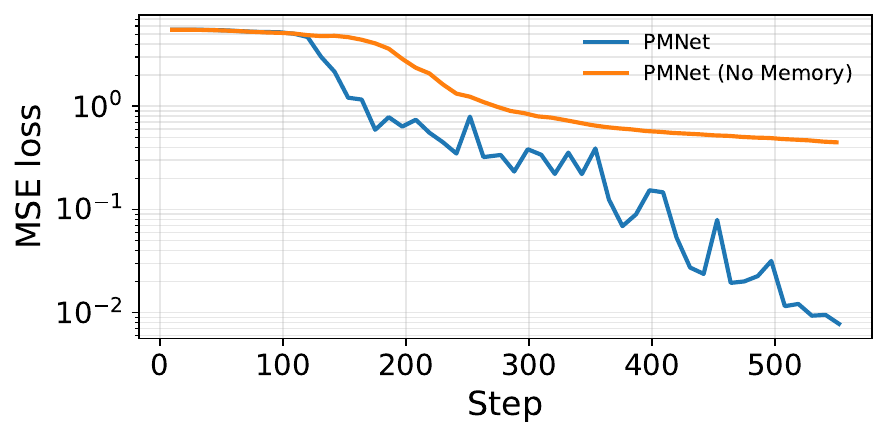}
        \caption{Evaluation Loss}
        \label{fig:copy_paste_eval_loss}
    \end{subfigure}
    \hfill
    \begin{subfigure}[t]{0.45\textwidth}
        \centering
        \includegraphics[width=\textwidth]{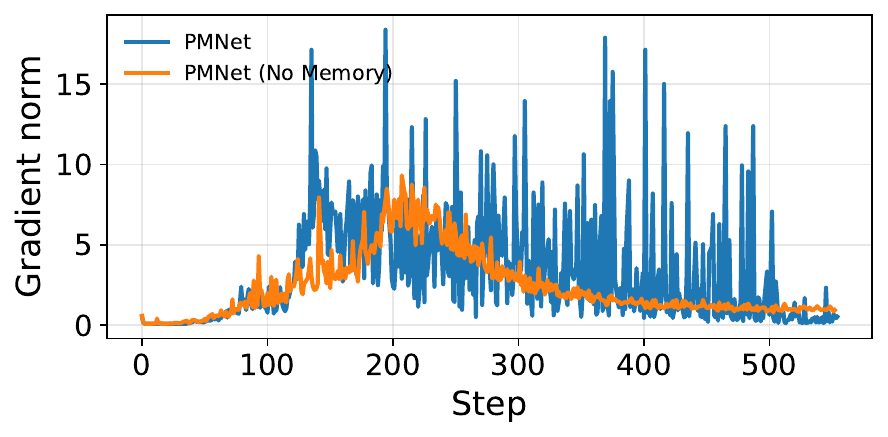}
        \caption{Gradient Norm}
        \label{fig:copy_paste_grad_norm}
    \end{subfigure}
    \caption{Training dynamics on the copy-paste task for PMNet (No memory) and PMNet. 
    PMNet (No Memory) was trained without memory modules, while PMNet was trained with memory modules.}
    \label{fig:copy_paste_training_dynamics}
\end{figure} 
\textbf{Experimental Setup:}
To verify whether the hierarchical phasor memory functions as a robust routing mechanism rather than a passive component, we evaluate PMNet on a synthetic Copy-Paste (Associative Recall) task.
This task acts as a mechanistic probe, requiring the model to memorize a random sequence of $N$ bytes and reproduce it exactly following a delimiter.
We utilized a 30M parameter PMNet consisting of 12 layers, each with a SWA size of 128.
Crucially, the global context in this configuration is managed by an expansive \textit{85-slot hierarchical memory tree} ($=\sum^{4}_{h=1}4^{h-1}$, corresponding to a depth of 4 and a branching factor of 4).
The local attention backbone imposes a strict theoretical receptive field limit of $12 \times 128 = 1,536$ bytes.
During training, the model was optimized to reconstruct sequences with \textit{variable lengths $N$ uniformly sampled between 10 and 1,024 bytes}. 
Since a Copy-Paste task of length $N$ requires a lookback distance of $2N$ (source + delimiter + target),
this training distribution forces the model to continuously adapt to temporal dependencies ranging from trivial short-range contexts up to long-range ones ($2N = 2,048$ bytes).
Consequently, any model relying solely on SWA should theoretically fail when $2N > 1,536$ (i.e., $N > 768$), making success on this task strictly contingent upon exact retrieval from the 85 explicit memory slots.

\textbf{Results:}
Figure \ref{fig:copy_paste_accuracy} illustrates the exact token accuracy.
The \textit{PMNet (No Memory)} baseline relies exclusively on SWA and maintains near 100\% accuracy up to $N=768$, perfectly aligning with its theoretical receptive field limit.
However, its performance catastrophically collapses to 0\% at $N=1,152$, where the source sequence resides entirely outside the SWA's reach.
In contrast, the full \textit{PMNet} maintains near 100\% accuracy even at $N=1,024$, demonstrating that the explicit memory module successfully routes and retrieves exact information across temporal distances that the local attention backbone cannot bridge.

\textbf{Gradient Dynamics Analysis:}
\textit{Why are Phasor Dynamics essential for BPTT?} The training gradient dynamics provide a clear mechanistic answer.
As shown in Figure \ref{fig:copy_paste_grad_norm}, the \textbf{PMNet (No Memory)} baseline exhibits a brief initial increase in gradient norm, corresponding to the early learning phase of local SWA.
However, it quickly stabilizes into a relatively calm, flat trajectory, indicating that the model has confined itself to local optimization and abandoned long-range dependencies it cannot resolve.
Conversely, the full \textbf{PMNet} exhibits a fascinating \textit{phase transition}.
Initially, the local SWA handles the optimization.
However, as the model is forced to solve dependencies beyond the SWA limit, it actively begins to route information into the global memory states.
This actuation triggers significant gradient spikes (up to $\approx 18$) due to the massive information exchange over long BPTT horizons.
Without structural constraints, these routing spikes would inevitably lead to catastrophic gradient divergence.
The successful convergence of the model despite these spikes provides definitive empirical proof that our stabilization mechanisms---the \textbf{unitary phasor updates} (preserving $|e^{i\theta}| \equiv 1$) and \textbf{segment-aware normalization}---function as the critical safety bounds that make deep NTMs trainable.

\subsection{Zero-Shot Extrapolation to Book-Length Sequences}
\label{sec:zero_shot_extrapolation}

\textbf{Training Setup:}
Having established the mechanistic stability of Phasor updates, we scale the model to fully unlock its data-hungry global routing capacity.
We trained a 119M parameter PMNet on approximately \textit{18.8 billion bytes} from \textit{FineWeb-Edu}, processing a context length of \textit{10,240 bytes} per sample. 
Despite this challenging long-context regime, PMNet maintained robust training stability without requiring complex regularization tricks often essential for traditional RNNs.
\textbf{Figure \ref{fig:training_dynamics}} visualizes the stable training dynamics in Appendix~\ref{sec:training_dynamics}.

\textbf{Extrapolation Results:}
We evaluated PMNet directly on the PG-19 test set in a zero-shot setting~\cite{rae2019compressive}.
Crucially, despite being trained on OOD data and possessing only 119M parameters, \textbf{PMNet (119M)} achieves performance parity with \textbf{MambaByte\_Arxiv (353M)}, a state-space model that is approximately three times larger and trained in-domain.
This underscores PMNet's exceptional parameter efficiency.
In contrast, \textbf{SmolLM (135M)} exhibits catastrophic divergence once the sequence length exceeds its pre-training context window, highlighting the inherent limitations of standard Transformers~\cite{allal2024smollm}.

\textbf{Ablation Study: Efficacy of Phasor Memory:}
We evaluated an ablated variant of the trained PMNet (119M) where the memory update mechanism was disabled during inference.
As illustrated by the red dashed line in Figure \ref{fig:long_context}, this results in a consistent degradation across the entire context length.
Although the quantitative gap appears modest, this is a known characteristic of language modeling, where prediction relies heavily on short-range patterns efficiently handled by SWA (Local Dominance).
Crucially, however, the full model achieves a strictly lower BPB than the ablation across \textit{all} token indices.
This consistent superiority proves that the phasor memory successfully captures the residual global context that remains inaccessible to local attention mechanisms.
The memory is not a placebo; it is a functionally integrated, actively retrieved knowledge base. 
\section{Limitations and Future Work}
While PMNet provides a mathematically sound solution to long-term explicit memory, has hardware-level optimization limitations.
In our pure PyTorch implementation, the segmented scan mechanism does not yet utilize highly fused custom CUDA kernels.
Consequently, the current wall-clock training time of PMNet is approximately $1.5\times$ that of a standard Vanilla Transformer and $3.0\times$ that of highly hardware-optimized Mamba.
We anticipate that developing dedicated custom kernels for the unitary phasor updates will largely close this throughput gap,
making PMNet practically competitive for extreme-scale foundation models.
Additionally, our empirical validations are currently constrained to the $100\text{M}$ parameter scale due to compute limits, leaving continuous scaling law studies to future work. While not yet positioned to immediately replace highly optimized, massive-scale fixed-context models, PMNet offers a theoretically grounded alternative.
It outlines a path toward architectures that do not merely process a sliding window of text, but actively curate and retain knowledge over extended temporal horizons.
We hope this work reignites interest in explicit memory systems as a fundamental component for the next generation of sequence modeling. 
\begin{ack}

\end{ack}

{
\small
\bibliographystyle{apalike}
\bibliography{ref}
}

\appendix

\newpage
\section{Notations}
\begin{table}[htb]
\centering
\caption{Summary of notations and tensor shapes used in PMNet.}
\label{tab:notations}
\begin{tabular}{rp{6.5cm}}
\toprule
\textbf{\small Symbol} & \textbf{\small Description} \\ \midrule
$l, t$ & Layer index, Time step \\
$h, p$ & Hierarchy level, Write period \\
$L$ & Total layers \\
$H$ & Memory hierarchy depth, L/P \\
$d$ & Dimension of hidden state \\
$d_m$ & Dimension of phasor memory vector \\
$N_{m}$ & Number of memory slots per group \\
$C_m$ & Total addressable memory capacity \\
$\mathbf{h}_{l,t}$ & Hidden state at layer $l$ and time step $t$ $\mathbb{R}^d$\\
$g_h$ & Memory group index at hierarchy level $h$ \\
$\mathbf{M}_t^{g_h}$ & Memory state of $g_h$ at time step $t$, $\mathbb{R}^{N_m \times d_m}$ \\
$\Phi(\mathbf{x})$ & $\Phi(x) = [\sin(\mathbf{x}), \cos(\mathbf{x})]$, $\mathbb{R}^{2d_m}$ \\
$\mathbf{E}^{g_h}$ & Addressable memory embeddings, $\mathbb{R}^{N_m \times d_m}$ \\ 
$w$ & Sliding window size of local attention \\
\bottomrule
\end{tabular}
\end{table} 
\newpage
\section{Parallel Segmented Scan with Segment-Aware Gradient Normalization}
While the inference of PMNet is inherently recurrent to ensure memory efficiency, the additive nature of the phasor update enables highly efficient parallelization during training. 
By exploiting the fact that phase updates are commutative, we reformulate the sequential memory accumulation as a Parallel Segmented Scan operation. Specifically, for a sequence of length $T$, we group tokens by their active memory group indices $g_h$ and perform a scan within each contiguous segment. 
This approach reduces the temporal complexity from $O(T)$ to $O(\log T)$, allowing the model to leverage the massive parallelism of modern GPUs.

\begin{algorithm}[ht]
  \caption{Parallel Segmented Scan with Segment-Aware Gradient Normalization}
  \label{alg:segmented_scan}
  \begin{algorithmic}[1]
    \State \textbf{Input:}
      Update grid $\Delta \mathbf{M} \in \mathbb{R}^{B \times S \times N \times M}$ \\
      Memory group indices $\mathcal{G} \in \mathbb{Z}^{B \times S}$ \\
      Total memory groups $N_{\text{grp}}$
    \State \Comment{\textbf{1. Routing \& Linearization}}
    \State Define total tokens $P \leftarrow B \cdot S$ and token indices $i \in \{0, \dots, P-1\}$
    \State Construct routing keys $\mathcal{K} \in \mathbb{Z}^P$:
    \State \quad $\mathcal{K}[i] \leftarrow \lfloor i/S \rfloor \cdot N_{\text{grp}} + \mathcal{G}_{\text{flat}}[i]$ \quad \Comment{Encode batch \& group}
    
    \State \Comment{Get permutation indices $\pi$ (corresponds to \texttt{perm} in code)}
    \State $\pi \leftarrow \text{StableArgsort}(\mathcal{K})$ 
    
    \State Permute keys and updates:
    \State \quad $\mathcal{K}_{\pi} \leftarrow \mathcal{K}[\pi], \quad \Delta \mathbf{M}_{\pi} \leftarrow \Delta \mathbf{M}_{\text{flat}}[\pi]$

    \State \Comment{\textbf{2. Boundary Detection}}
    \State Detect segment boundaries:
    \State \quad $b_i \leftarrow \mathbb{I}(\mathcal{K}_{\pi}[i] \neq \mathcal{K}_{\pi}[i-1]) \quad \text{for } i > 0, \quad b_0 \leftarrow 1$
    \State Define start indices $\mathcal{S}_{\text{start}} \leftarrow \{ i \mid b_i = 1 \}$ and end indices $\mathcal{S}_{\text{end}}$
    \State Compute segment lengths $\mathbf{L} \in \mathbb{Z}^K$ using boundary differences.

    \State \Comment{\textbf{3. Gradient Scaling (Training Only)}}
    \If{training}
      \State Broadcast lengths to token level: $\mathbf{L}_{\text{sorted}} \leftarrow \text{Repeat}(\mathbf{L})$
      \State Map lengths to temporal order: $\mathbf{L}_{\text{flat}}[\pi] \leftarrow \mathbf{L}_{\text{sorted}}$
      \State Compute scale factor: $\boldsymbol{\sigma} \leftarrow \sqrt{\max(1, \mathbf{L}_{\text{flat}})}$
      \State Apply Straight-Through Estimator to input grid:
      \State \quad $\Delta \mathbf{M} \leftarrow \frac{\Delta \mathbf{M}}{\boldsymbol{\sigma}} + \text{stopgrad}\left( \Delta \mathbf{M} - \frac{\Delta \mathbf{M}}{\boldsymbol{\sigma}} \right)$
    \EndIf

    \State \Comment{\textbf{4. Parallel Segmented Scan}}
    \State Sort updates: $\Delta \mathbf{M}_{\pi} \leftarrow \Delta \mathbf{M}[\pi]$
    \State Compute global prefix sums: $\mathbf{C} \leftarrow \text{PrefixSum}(\Delta \mathbf{M}_{\pi}, \text{dim}=0)$
    \State Extract offsets at segment starts: $\mathbf{O} \leftarrow \mathbf{0} \oplus \mathbf{C}[:\!-1]$, gather at $\mathcal{S}_{\text{start}}$
    \State Compute local segment states: 
    \State \quad $\mathbf{Z}_{\pi} \leftarrow \mathbf{C} - \text{Broadcast}(\mathbf{O})$

    \State \Comment{\textbf{5. Restoration}}
    \State Invert permutation to restore temporal order:
    \State \quad $\mathbf{R}_{\text{flat}}[\pi] \leftarrow \mathbf{Z}_{\pi}$
    \State Apply phase projection: $\mathbf{R} \leftarrow \mathbf{R}_{\text{flat}} \pmod{2\pi}$
    \State Reshape $\mathbf{R}$ to $B \times S \times N \times M$
    \State Return restored updates $\mathbf{R}$
  \end{algorithmic}
\end{algorithm} 
\newpage
\section{Training Curves for Zero-Shot Extrapolation to Book-Length Sequences}
\label{sec:training_dynamics}
\begin{figure}[htbp]
    \centering
    \begin{subfigure}[t]{0.48\textwidth}
        \centering
        \includegraphics[width=\textwidth]{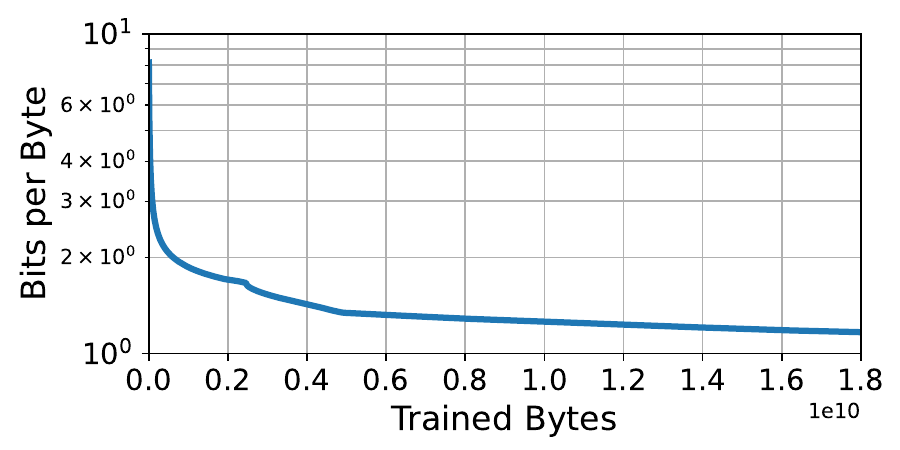}
        \caption{Training Curve}
        \label{fig:training_curve}
    \end{subfigure}
    \hfill
    \begin{subfigure}[t]{0.48\textwidth}
        \centering
        \includegraphics[width=\textwidth]{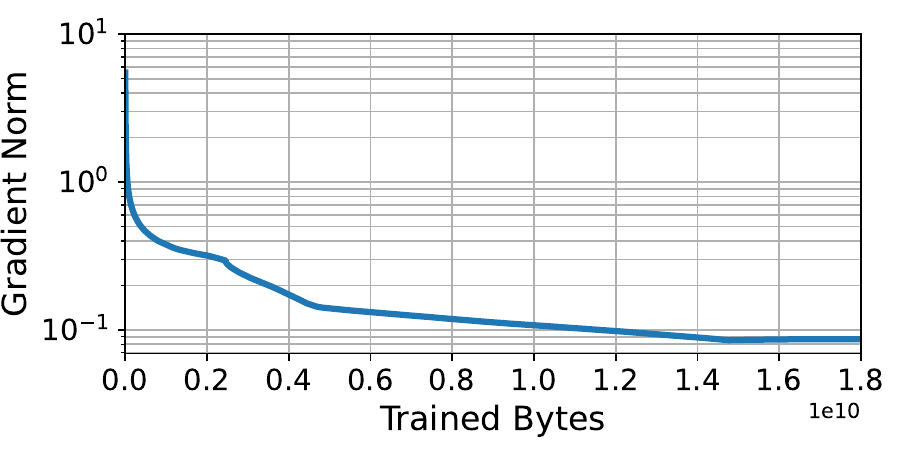}
        \caption{Gradient Norm}
        \label{fig:grad_norm}
    \end{subfigure}
    \caption{(1) \textbf{Monotonic Power-Law Convergence:} PMNet follows a smooth power-law scaling curve, indicating that the hierarchical memory is effectively compressing information rather than accumulating noise. \\
    (2) \textbf{Bounded Gradient Norms:} As shown in Figure \ref{fig:grad_norm}, gradient norms remain strictly bounded throughout the entire 18.8B training process, empirically validating that the exploding gradient problem has been structurally eliminated.}
    \label{fig:training_dynamics}
\end{figure} 
\newpage
\section{Structural Stability and Addressing Drift Analysis}
\begin{figure}[htbp]
    \centering
    \includegraphics[width=.5\textwidth]{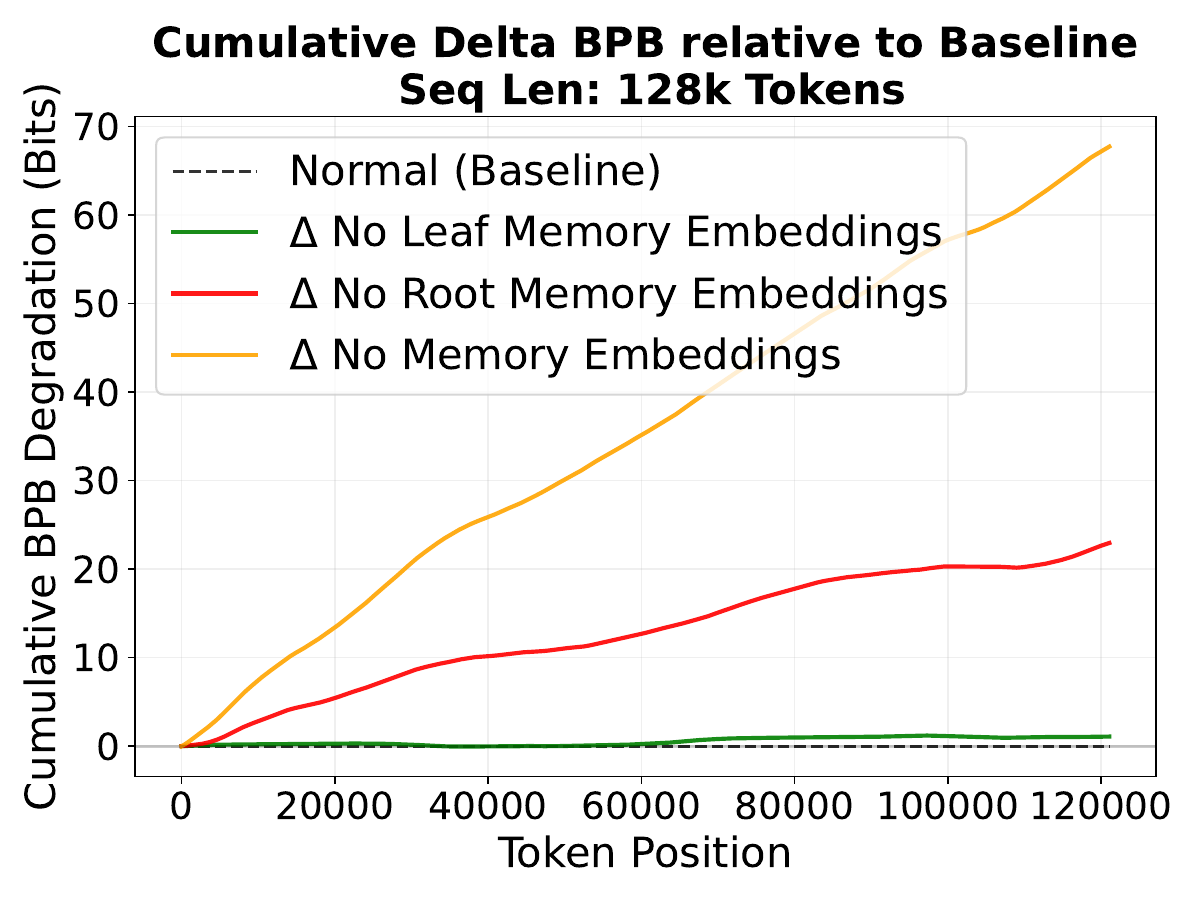} 
    \caption{
    \textbf{Structural Stability via Hierarchical Embeddings.} 
    Cumulative Delta BPB over a 128k sequence (PG-19) under zero-shot ablation. Removing all embeddings (\textbf{Orange}) causes linear addressing drift. The stark contrast between ablating the root (\textbf{Red}) and leaf (\textbf{Green}) validates our design: roots form the critical addressing schema, while leaves offer scalable, fine-grained detail storage.}
    \label{fig:memory_embedding}
\end{figure} 
A critical challenge in explicit memory architectures, such as NTM, is addressing drift—the tendency for the model to lose track of memory locations over long sequences due to the accumulation of numerical errors and the lack of stable references.
To verify whether PMNet's Hierarchical Memory with Learnable Embeddings effectively resolves this issue, we conducted a zero-shot ablation study.
We selectively zeroed out specific memory embeddings during inference on a 128k token sequence on PG-19 dataset and measured the Cumulative Delta BPB relative to the full model (Normal).
This method isolates the structural contribution of each memory component without confounding optimization factors.

\paragraph{Preventing Addressing Drift}
As shown in Figure \ref{fig:memory_embedding} (Orange Line), completely removing the memory embeddings (mimicking a traditional drift-prone NTM) results in a linear accumulation of error over time. 
The learnable embeddings act as static semantic pegs, allowing the model to offload addressing stability to the fixed hierarchy while reserving recurrent capacity for dynamic content.

\paragraph{Validating the Hierarchy (Root vs. Leaf)}
The ablation results reveal a clear functional distinction between hierarchical levels. (Figure \ref{fig:memory_embedding}, Red vs. Green) \\
\textbf{No Root (Red):} Zeroing out the root memory embeddings causes a rapid and steep increase in cumulative loss.
This indicates that the root nodes function as the addressing backbone or schema, directing the global flow of information. \\
\textbf{No Leaf (Green):} In contrast, removing leaf memory embeddings results in negligible performance degradation.
This suggests that leaf nodes handle fine-grained, redundant details with high plasticity. 

This disparity supports our hypothesis of a schema-detail dichotomy: the model relies on the root for structural stability while using leaves for high-capacity detail storage.
Critically, the minimal impact of leaf ablation implies scalability; we can expand the memory capacity at the leaf level without destabilizing the core addressing mechanism.

\newpage

\end{document}